\documentclass[runningheads]{llncs}
\usepackage{amsmath}
\usepackage{booktabs} 
\usepackage{caption} 
\usepackage{subcaption} 
\usepackage{graphicx}
\usepackage{pgfplots}
\usepackage[utf8]{inputenc}
\usepackage{multicol}
\usepackage{hyperref}
\usepackage[inline]{enumitem} 
\definecolor{blue}{HTML}{1F77B4}
\definecolor{orange}{HTML}{FF7F0E}
\definecolor{green}{HTML}{2CA02C}

\pgfplotsset{compat=1.14}

\begin{document}
\title{Multilingual Central Repository: a Cross-lingual Framework for Developing Wordnets}
\titlerunning{Multilingual Central Repository}
%
\author{Xavier Gómez Guinovart\inst{1} \and
Itziar Gonzalez-Dios\inst{2} \and \\
Antoni Oliver\inst{3}\and
German Rigau\inst{2}}


%
\authorrunning{Gómez Guinovart et al.}
%

\institute{Seminario de Lingüística Informática (SLI), 	Universidade de Vigo \\
\email{xgg@uvigo.gal}\\ \and
	Ixa group, HiTZ center,University of the Basque Country (UPV/EHU)\\
\email{\{itziar.gonzalezd,german.rigau\}@ehu.eus}\and
Universitat Oberta de Catalunya\\
\email{aoliverg@uoc.edu}
}
\maketitle              
\begin{abstract}
Language resources are necessary for language processing, but building them is costly, involves many researches from different areas and needs constant updating. In this paper, we describe  the cross-lingual framework used for developing the Multilingual Central Repository (MCR), a multilingual knowledge base that includes wordnets of Basque, Catalan, English, Galician, Portuguese, Spanish and the following ontologies: Base Concepts,  Top Ontology, WordNet Domains and Suggested Upper Merged Ontology. We present the story of MCR, its state in 2017 and the developed tools. 

\end{abstract}
\section{Introduction} 

Building large and rich knowledge bases  and language resources is a very costly effort which involves large research groups for long periods of development. For instance, hundreds of person-years have been invested in the development of wordnets for various languages \cite{fellbaum98,Vossen:1998,Tufics04,Robkop10}. In the case of the English WordNet, in more than ten years of manual construction (from 1995 to 2006, that is, from version 1.5 to 3.0), WordNet grew from 103,445 to 235,402 semantic relations\footnote{Symmetric relations are counted only once.}, which represents a growth of around one thousand new relations per month. This is also the case of the Iberian wordnets developed and integrated into the Multilingual Central Repository\footnote{\url{http://adimen.si.ehu.es/web/MCR}} \cite{gonzalez2012multilingual}. 

This paper describes the cross-lingual framework used for developing the Multilingual Central Repository (MCR). Currently, the MCR uses WordNet 3.0 as Interlingual-Index (ILI) and integrates in the same EuroWordNet framework wordnets from six different languages: English, Spanish, Catalan, Basque, Galician and Portuguese. In order to provide ontological coherence to all the integrated wordnets, the MCR has also been enriched with a disparate set of ontologies: Base Concepts,  Top Ontology, WordNet Domains and Suggested Upper Merged Ontology.  
The whole content of the MCR is freely available under  the original WordNet license for the English and CC BY 3.0 for the others.

\section{A brief history of the MCR} 

The Multilingual Central Repository (MCR)\footnote{http://adimen.si.ehu.es/web/MCR} \cite{atserias04}, which follows the model proposed by the EuroWordNet project (LE-2 4003) \cite{Vossen:1998}, is the result of the MEANING project (IST-2001-34460) \cite{rigau02}, as well as projects KNOW (TIN2006-15049-C03)\footnote{http://ixa.si.ehu.es/know} \cite{agirre09}, KNOW2 (TIN2009-14715-C04)\footnote{http://ixa.si.ehu.es/know2} and several complementary actions associated  to the KNOW$^2$ project. The original MCR was aligned to the 1.6 version of WordNet. In the framework of the KNOW$^2$ project, we decided to upgrade the MCR to be aligned to a most recent version of WordNet.

The previous version of the MCR was aligned to the English 1.6 WordNet version, it also integrated the eXtended WordNet project \cite{mihalcea01}, large collections of selectional preferences acquired from SemCor \cite{AgirreMartinez01} and different sets of named entities \cite{AlfonsecaManandhar02}. It was also enriched with semantic and ontological properties as  Base Level Concepts, \cite{izquierdo2007BLCs}, Top Ontology \cite{AAC08},  SUMO \cite{Niles+Pease'01} or WordNet Domains\footnote{http://wndomains.fbk.eu/} \cite{magnini00,bentivogli2004revising}.

The new MCR  integrates wordnets of six different languages, including English, Spanish, Catalan, Basque, Galician and Portuguese. This paper presents the work carried out to upgrade the MCR to new versions of these resources. By using technology to automatically align wordnets \cite{daude03}, we have been able to transport knowledge from different WordNet versions. Thus, we can maintain the compatibility between all the knowledge bases that use a particular version of WordNet as a sense repository. 

However, most of the ontological knowledge has not been directly ported from the previous version of the MCR. Furthermore, WordNet Domains was generated semi-automatically and has never been verified completely. Additionally, it was aligned to WordNet 1.6. 

Thus, one goal of this work is the automatic construction of a new semantic resource derived from WordNet Domains and aligned to WordNet 3.0.

Exactly and briefly, the most important  changes introduced in this  new version of the MCR are: 

\begin{itemize}
\item The  Portuguese  WordNet  (PULO) developed  at the University of Minho  

\item New  variants  for  Spanish,  Catalan,  Basque  and  Galician wordnets

\item New encoding for the MCR relations 

\item  A new version of Base Level Concepts (BLC)

\item  Correcting minor mistakes. 
\end{itemize}

\subsection{Languages in the MCR}

As already mentioned, the MCR includes wordnets for six languages: English, Spanish, Catalan, Portuguese, Galician and Basque. The English wordnet is included directly from the PWN and it serves as a reference for the synsets.

Four of the remaining languages, namely Spanish, Catalan, Portuguese and Galician, are Romance languages derived from Latin and share a lot of linguistic characteristics. On the other hand, Basque is an isolated language and is unrelated to any known living language. By contact with neighbor languages, Basque has adopted words from Latin, Spanish and Gascon, but these words have suffered changes due to Basque phonology and grammar.

Regarding the creation of the wordnets for these languages, the  first  version  of  Spanish  WordNet (1.5) \cite{atserias97,Farreres1998using} was build during the EuroWordNet project (Vossen, 1998). It was built following the  expand model: namely, WordNet synsets were translated into equivalent synsets in Spanish exploiting several Spanish-English bilingual dictionaries and following an automatic method. Then, it was upgraded to versions 1.6 \cite{atserias04}  and 3.0 \cite{gonzalez2012multilingual}.  To build the Catalan Wordnet \cite{benitez1998methods} the  methodology which was  applied  to  Spanish  WordNet was followed. The Basque wordnet was created both following manual and semi-automatic approaches \cite{pociello2011methodology} in the 1.6 versions and then upgraded.  Similar to Basque the Galician wordnet \cite{galnet}  was built following  manual and semi-automatic approaches but in the 3.0 version. Finally, the PULO Portuguese wordnet \cite{simoes2014bootstrapping} was bootstrapped from dictionaries and  the English, Galician and Spanish wordnets.

\subsection{Ontologies in the  MCR}

In the MCR there are some ontologies and hierarchies. Following, we present the main features of them.

Basic Level Concepts (BLC) are those concepts that are frequent and salient; they are neither overly general nor too specific. They try i)    to represent as many concepts as possible (abstract concepts) and ii)     to represent as many distinctive features as possible (concrete concepts).   

The EuroWordNet top-ontology (TO) is a lattice structure of 63 features that can be combined in feature combinations. It was designed as  an independent hierarchy of features  for clustering, comparing and exchanging concepts across languages in the EuroWordNet Project. Its first level is divided in three parts:  {\it 1stOrderEntity} for physical things, {\it 2ndOrderEntity} for events, states and properties and {\it  3rdOrderEntity} for unobservable entities. 

The Suggested Upper Merged Ontology (SUMO) is an ontology is a standard upper ontology that  promotes data interoperability,  information search and retrieval,  automated inference and natural language processing.  SUMO provides also  definitions for general purpose terms and consists of a set of concepts, relations, and axioms.  Adimen-SUMO \cite{ALR12} is a first-order logic (FOL) ontology obtained by means of a suitable transformation of most of the knowledge (around 88~\% of the axioms) in the {\it top} and {\it middle} levels of SUMO. 

WordNet domains is a lexical resource where synsets have been semi-automatically annotated with one or more domain labels from a set of 165 hierarchically organized labels. The aim of this resource is to  reduce  the  level  of  polysemy  of  the  senses, grouping those senses that belong to the same domain.

\section{MCR architecture} 

The current MCR is stored on a relational database consisting of 46 tables: 1 for the ILI, 30 for the WNs (5 per language), 2 for WordNet Domains, 2 for SUMO, 1 for BLCs, 3 for Top Ontology, 2 for the marks and 5 for define values  (relations,  groups  of  relations,  colors,  counters  and  semantic  files).  The most important tables are: 

\begin{itemize}
\item {\bf wei\_ili\_record} contains  the  identifier  of  the  ILI
\item {\bf wei\_xxx-30\_to\_ili}, {\bf wei\_xxx-30\_relation}, {\bf wei\_xxx-30\_synset}, {\bf wei\_xxx-30\_variant} and {\bf wei\_xxx-30\_examples} contain each  wordnet: connection to the ILI, the relations, the synsets, the variants and the examples. The xxx indicates the three letter code its language has, which are namely eng, spa, por, gal, cat and eus. 
\item {\bf  wei\_ili\_to\_blc}  contains the links of ILI to its BLC’s ILI.
\item {\bf wei\_sumo\_relations} and {\bf wei\_ili\_to\_sumo} contain the SUMO relations and the  links of SUMO label to an ILI.
\item {\bf  wei\_to\_relations}, {\bf  wei\_ili\_to\_to} and {\bf wei\_to\_record} indicate the TO hierarchy, the links to the ILIs  and the TO labels respectively.
\end{itemize}

\begin{figure}[h]
\begin{center}
\includegraphics[scale=0.7]{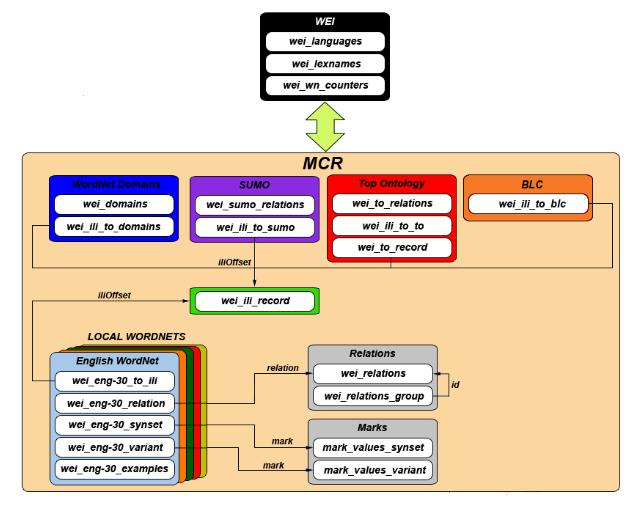}
\end{center}
\caption{Database structure of the MCR and WEI}
\label{fig:mcr-archi}
\end{figure}

In Figure \ref{fig:mcr-archi} we show the architecture of the MCR and the connection to the Web EuroWordnet Interface (WEI) that we will present in the following section.

\section{MCR tools} 
In this section we present the tools that have been developed in relation to the MCR.

\subsection{Interfaces: WEI and Galnet}

The MCR can be consulted in two different interfaces: the one developed in the Web EuroWordnet Interface (WEI) and the Galnet interface.

The WEI interface was developed during the EuroWordNet project and it has both  consult and edit mode. For this release the most changes we have made  are:
\begin{itemize}
\item add links to BabelNet and OWN webpages,   
\item upgrade  visualization  of  hierarchies,
\item  add a new version of AdimenSUMO, and
\item changed some buttons in edit mode.
\end{itemize}

The Galnet interface\footnote{\url{http://sli.uvigo.gal/galnet/}} was  designed to query Galnet, the Galician version of WordNet which is part of the MCR, extends the MCR WEI functionalities by providing
\begin{itemize}
\item different types of navigation through domain hierarchies and ontologies,
\item an interactive tree-based visualization of synsets by their semantic relations,
\item temporal values and sentiment scores for synsets from TempoWordNet\footnote{\url{https://tempowordnet.greyc.fr}}, SentiWordNet 3.0\footnote{\url{http://sentiwordnet.isti.cnr.it}} and ML-SentiCon\footnote{\url{http://sentiwordnet.isti.cnr.it}},
\item images associated to synsets from ImageNet\footnote{\url{http://www.image-net.org}}
\item a tool called Termonet specifically designed for the extraction of lexical-semantic fields,
\item a terminology-oriented semantic categorization based on epinonyms,
\item and a new presentation of information associated with synsets in Linked Open Data format (RDF Galnet).
\end{itemize}

\subsection{Termonet and epinonyms}

To explore terminology,   the possibilities of building a term-oriented hierarchical structure from all the set of the WordNet synsets were explored. In addition, a method was sought to verify the empirical occurrence of the concepts in specialized corpora.

To that end,  Termonet  enables the extraction of domain-specific variants from WordNet and provides a query form that allows selecting a synset from the lexical-semantic network and extracting related terms according to the semantic relations defined in the configuration. Although Termonet allows extraction from any WordNet synset, due to its terminological nature, the application always tries to suggest the closest noun variants when initiating a search from a non-noun synset.  Termonet's features rely on two basic resources: a wordnet and a corpus of specialized language. 

Research on Termonet has lead us to a new semantic categorization of WordNet devised to exploit the terminological implications of the relations between synsets. The adopted approach was based on tracing a path in the opposite direction to that used by Termonet to explore a domain from a synset, so each synset finds its way through the relations to an epinonym noun synset representing the semantic domain in which to be included automatically. Thus, an epinonym is a noun synset representing the category of the semantic domain to which other synsets will be automatically assigned by algorithms that will evaluate their proximity from a terminological point of view through the cognitive processing of the lexical semantic relations in the network \cite{galnet}.

\subsection{RDF versions}

Galnet interface also provides all its contents as RDF resources through a SPARQL endpoint\footnote{\url{http://sli.uvigo.gal/sparql/}}  with free public access for users to explore the data using SPARQL queries \cite{dbpedia2016}. 

The RDF Galnet monolingual dictionaries conform to the Lemon model.\footnote{\url{http://lemon-model.net}} The Galnet synsets are aligned with Princeton's WordNet synsets version 3.1, with Princeton's WordNet synsets version 3.0 in lemonUby\footnote{\url{http://lemon-model.net/lexica/uby/}}   and with the Interlingual Index (ILI). In many cases, Princeton's WordNet also provides the alignment with a corresponding synset in lemonUby version 3.0. However, the alignment in RDF Galnet offers correspondences between all MCR synsets from version 3.0 and version 3.1 or lemonUby ones.

The RDF Galnet internal ontology is based on the ontology that uses the RDF Princeton Wordnet 3.1, revised and adapted to the EuroWordNet framework followed by the MCR project. Moreover, all the ontologies linked to the Galnet synsets were converted to the RDF data model: Adimen-SUMO, Top Ontology, WordNet Domains and Epinonyms. Turtle files with data corresponding to the latest public release of the dataset and related ontologies can also be downloaded from Galnet site.\footnote{\url{http://sli.uvigo.gal/download/SLI_Galnet/}}


\subsection{The WN-Toolkit}

The expansion of the Spanish, Catalan and Galician wordnets in the MCR has been partially performed using the WN-Toolkit\footnote{\url{http://sourceforge.net/projects/wn-toolkit/}} \cite{oliver2014wn}. This is a set of programs and data sets for the automatic creation of wordnets following the expand model, that is, by the translation of English variants from synsets of the PWN. The toolkit implements several strategies:

\begin{itemize}
\item Dictionary based methodology using bilingual dictionaries. The direct translation of English variants can be only done for monosemic variants, that is, variants assigned to a single synset. For polysemic variants several strategies have been developed to use the definitions in WordNet and the lexical resource in order to select the correct alignment. Within the toolkit, several alignments between PWN and some free lexical resources (namely Wikipedia, Wiktionary and Omegawiki) have been published.

\item Using old versions of BabelNet \cite{navigli2010babelnet}, where relations between PWN and Wikipedia are provided.

\item Parallel corpus based methodologies. For this methodology the corpus should be semantically tagged with PWN synsets. As these corpora are not easily available, two strategies have been used:

\begin{itemize}
\item Machine translation of sense-tagged corpora
\item By automatic sense-tagging of the English part of available English-target language parallel corpora.
\end{itemize}
\end{itemize}

The WN-Toolkit has been successfully use to enlarge other wordnets, as for example the Croatian wordnet \cite{oliver2015enlarging}. 

The future development of this toolkit is oriented to the creation of reliable alignments between wordnets and Wikipedia, Wiktionary and Omegawiki. These alignments will allow the creation of new wordnets and the expansion of existing ones.

\subsection{UKB: Graph Based Word Sense Disambiguation and Similarity}

One of the most well-known  Word Sense Disambiguation (WSD) tool is UKB \cite{agirre2018UKB}, which is  is a collection of programs for performing graph-based WDS. Using  pre-existing knowledge bases (KB) like wordnets in MCR,  UKB applies random walks, e.g. Personalized PageRank, on the KB graph to rank the vertices according to the given context. UKB can be used together with the IXA-pipes \cite{agerri2014ixapipes}, a  modular set of NLP tools (or pipes)  for several languages, in order to perform WSD.

\section{Current state of the MCR}

The MCR is actively maintained and the wordnets in the MCR are still under continuous development in order to enlarge and improve them.   The latest distribution of MCR was done in 2016 can be downloaded from \url{http://adimen.si.ehu.es/web/MCR/}, but regular updates can be consulted in the interface. 



Following we show the statistics for the Spanish, Catalan, Portuguese, Galician and Basque wordnets included in the MCR in the latest distribution.  In Table \ref{tab:mcr2016-stats} we present the main statistics for all the languages, where Core \% is the percentage of core synsets covered;     CILI \% is the percentage of synsets linked to CILI; and     Def and Ex \% are the percentages of synsets       with definitions and examples respectively.


\begin{table*}[!h]
\centering 
  \begin{tabular}{lrrrrr}
 & Spa & Cat & Por & Gal & Eus \\   \hline
  Synsets &   38,512 &   45,826 & 15,608 & 40,975 &  29,413 \\ 
 Words &   37,203  &    47,598  &  8,471  &   50,702  &  26,390 \\ 
Forms &  37,203  & 47,598  &   8,471  &   50,702  &  26,390 \\ 
  Senses & 57,764 & 70,622 &  21,244 &   64,338 & 48,934 \\ 
 Core \% &   76.0 &  81.0 &  64.3 &   74.7 &  70.5 \\ 
  CILI \% &100.0 & 100.0 &  100.0 &  100.0 & 100.0 \\ 
 Def \% &  0.0 &    0.0 &   0.0 &  21.3 & 0.0  \\ 
 Ex \% & 0.0   0.0 &   0.0 & 0.0 & 0.0   & 0.0 \\ \hline
  \end{tabular}
   \caption{Main Statistics for MCR 3.0 (2016)}
  \label{tab:mcr2016-stats}
\end{table*}

Following, we show the PoS  statistics for each of the wordnets in Table \ref{tab:mcr_all-3.0-2016:stats:pos}.



      

\begin{table*}[!h]
\centering
  
 \begin{tabular}{llrrrrrr}
         {\bf Lang} & {\bf PoS} & {\bf Synsets} & {\bf  \% }  & {\bf Words} &  {\bf  \% }  & {\bf Senses} & {\bf  \% }   \\
      
 \hline 
 
  {\bf Spa} &
    Noun &
    26,404 & 
    68.6 & 
    28,647 &
    77.0 &
    38,917 &
    67.4 \\
  &  
    Verb &
    6,251 & 
    16.2 & 
    4,354 &
    11.7 &
    10,829 &
    18.7 \\
  &  
    Adjective &
    5,180 & 
    13.5 & 
    3,289 &
    8.8 &
    6,967 &
    12.1 \\
 &   
    Adverb &
    677 & 
    1.8 & 
    913 &
    2.5 &
    1,051 &
    1.8 \\  \hline
    
   {\bf Cat}   &
        Noun &
    36,253 & 
    79.1 & 
    38,733 &
    81.4 &
    51,364 &
    72.7 \\
    &
    Verb &
    5,424 & 
    11.8 & 
    4,633 &
    9.7 &
    11,577 &
    16.4 \\
    &
    Adjective &
    4,148 & 
    9.1 & 
    4,230 &
    8.9 &
    7,679 &
    10.9 \\
    &
    Adverb &
    1 & 
    0.0 & 
    2 &
    0.0 &
    2 &
    0.0 \\  \hline
    
    {\bf Por}   &
     Noun &
    8,750 & 
    56.1 & 
    5,057 &
    59.7 &
    11,760 &
    55.4 \\
      &
    Verb &
    3,200 & 
    20.5 & 
    1,418 &
    16.7 &
    4,840 &
    22.8 \\
      &
    Adjective &
    3,175 & 
    20.3 & 
    1,582 &
    18.7 &
    4,118 &
    19.4 \\
      &
    Adverb &
    483 & 
    3.1 & 
    414 &
    4.9 &
    526 &
    2.5 \\ \hline
    
     {\bf Gal}   & 
      Noun &
    31,163 & 
    76.1 & 
    39,617 &
    78.1 &
    46,634 &
    72.5 \\
    & 
    Verb &
    3,212 & 
    7.8 & 
    4,079 &
    8.0 &
    7,237 &
    11.2 \\
    & 
    Adjective &
    5,526 & 
    13.5 & 
    5,801 &
    11.4 &
    8,823 &
    13.7 \\
    & 
    Adverb &
    1,074 & 
    2.6 & 
    1,123 &
    2.2 &
    1,644 &
    2.6 \\  \hline
    
    {\bf Eus}   & 
        Noun &
    25,938 & 
    88.2 & 
    22,877 &
    86.7 &
    39,535 &
    80.8 \\
    & 
    Verb &
    3,364 & 
    11.4 & 
    3,456 &
    13.1 &
    9,251 &
    18.9 \\
    & 
    Adjective &
    111 & 
    0.4 & 
    57 &
    0.2 &
    148 &
    0.3 \\
          \hline 
          \end{tabular}
          \caption{PoS Statistics for wordnets in  MCR 3.0 (2016)}
  \label{tab:mcr_all-3.0-2016:stats:pos} 
   \end{table*} 



\section{Concluding Remarks and Future Plans}


In this paper we have presented the Multilingual Central Repository (MCR), a large scale knowledge base on continuous development since 2002. Currently, the  MCR includes wordnets of six different languages (English, Spanish, Catalan, Basque, Galician and Portuguese) and ontological knowledge provided by the Basic Level Concepts, the EuroWordNet top-ontology,  SUMO and WordNet domains.

From now on, we plan  to apply advanced deep learning techniques for building automatically large-scale wordnets from scratch from any language and domain. We want to explore novel deep learning approaches and methods for acquiring general and specialized large-scale lexical knowledge from textual corpora. The domains that will be specially targeted are eHealth, eLearning, eTourism and eJustice.  This task covers the exploitation of word and sense embeddings for creating new synsets in order to include new concepts for all the MCR languages based on the Collaborative ILI. Additionally, existing glosses will be also translated by means of supervised and unsupervised Neural Machine Translation techniques.

Moreover, we will integrate the semantic categorization based on epinonyms into the MCR and the Predicate Matrix \cite{de2016predicate}, a new lexical resource resulting from the integration of multiple sources of predicate information including FrameNet \cite{baker1998berkeley}, VerbNet \cite{schuler2005verbnet}, PropBank \cite{palmer2005proposition}, WordNet \cite{fellbaum98}, Basque Verb Index \cite{estarrona2020corpus} and ESO \cite{segers2015eso}.

Finally, by leveraging information from textual data and existing knowledge bases and ontologies, we would like to address a very difficult challenge in text mining: checking the veracity of answers based on external sources such as the MCR.

\section*{Acknowledgments}
This research has been carried out thanks to the project DeepReading (RTI2018-096846-B-C21) supported by the Ministry of Science, Innovation and Universities of the Spanish Government. 

%
%
\bibliographystyle{splncs04}
\bibliography{mcr}
\end{document}